# A WL-SPPIM Semantic Model for Document Classification

First A. Ming Li, Second B. Peilun Xiao, and Third C. Ju Zhang

**Abstract**—In this paper, we explore SPPIM-based text classification method, and the experiment reveals that the SPPIM method is equal to or even superior than SGNS method in text classification task on three international and standard text datasets, namely 20newsgroups, Reuters52 and WebKB. Comparing to SGNS, although SPPMI provides a better solution, it is not necessarily better than SGNS in text classification tasks.. Based on our analysis, SGNS takes into the consideration of weight calculation during decomposition process, so it has better performance than SPPIM in some standard datasets. Inspired by this, we propose a WL-SPPIM semantic model based on SPPIM model, and experiment shows that WL-SPPIM approach has better classification and higher scalability in the text classification task compared with LDA, SGNS and SPPIM approaches.

——————————— ◆ ———————————

## 1 INTRODUCTION

Distribution of semantic vectors is widely used in text semantic expression, including text classification, text clustering, semantic retrieval, automatic question and answer, dictionary generation, semantic disambiguation, query expansion, text advertisements and machine translation, especially for measuring semantic relevance[1, 2]. We divided the DSMs into two categories, one we called count-based models, many traditional DSMs belong to this category, the other category we call prediction-based models, which are based on neural embedding

Among the traditional count-based models the best know is Latent Semantic Analysis. LSA is a low dimensional semantic space for texts, and LSA derive the document vector by the use of co-occurrence information between words[3]. More recently, LDA has received more and more extensive attention, as a semantic model of DSMs[4, 5]. LDA is a three-layer Bayesian probability model proposed by Blei et al in 2003, which contains the three layers structure of document, topic and word. Document to topic subjects to Dirichlet distribution, topic to word subjects to polynomial distribution. LDA semantic model usually shows very good performance on NLP tasks, partly because it projects the document into a low-dimensional topic semantic space. Word frequencies determine the topic of calculated and deducted by LDA largely, which leads less frequent but important topics can not be effectively calculated.

Pointwise mutual information(PMI) has been extensively uesd as count-based model in distributional semantic models. Pointwise Mutual Information (PMI) is the popular word co-occurrence based measure[6, 7]. PMI has a well-known tendency that it calculates too high scores for low frequency words[2]. To solve the limitation fo PMI, many variants of it have been proposed, PPMI is the simple one of the variants, in which all PMI values that below zero will be set to zero[8]. Bullinaria and Levydem demonstrate that PPMI outperforms various other weighting methods，when measuring semantic similarity through word-context matrices.[9].

Traditional distributional Semantic Models achieve considerable effective effects on various NLP tasks, including semantic relevance and text classification.

The last few years have seen the development of prediction-based neural embedding models in which words are embedded into a low dimensional space. Word embedding models can efficiently learn word vector from a large number of unstructured text, and can effectively reflect the syntactic or semantic relations between words[10]. The initial pioneering research work was started by Bengio and his colleagues, who generated word vectors in the study of the neural language model[11] and a number of subsequent research work including various word embedding models and efficient learning algorithms[12-16]. In particular, We notice a conclusion that the SGNS(skip-gram with negative-sampling)model which is efficient to provides competitive results on various NLP tasks, which is concluded in a sequence of papers by Mikolov and colleagues[14, 17]. The SGNS model maximizes the conditional probability of the observed contexts given the current word when scanning through the corpus, however it is not clear what information the embedding vectors really convey, so prediction-based neural embeddings are considered opaque.

The researchers have different conclusions on the various types of distributed semantic models applying to the performance of NLP tasks, some of these conclusions

----

- F. A. Author is with High performance computing application R&D Center, Chongqing Institute of Green and Intelligent Technology, Chinese Academy of Sciences, Chongqing, China & University of Chinese Academy of Sciences, Beijing, China & Center for Speech and Language Technology,Research Institute of Information Technology, Tsinghua University,Beijing, China; E-mail: liming@cigit.ac.cn.
- S.B. Author is an undergraduate in Beijing University of Posts and Telecommunications. This work was done when he was a visiting student at Tsinghua University; Email:xiaopli@cslt.riit.tsinghua.edu.cn.
- T. C. Author is with with High performance computing application R&D Center, Chongqing Institute of Green and Intelligent Technology, Chinese Academy of Sciences Chongqing, China; E-mail: Zhangju@cigit.ac.cn.

are the same, some are the opposite, some conclusions depend on the distributed semantic models application of specific NLP tasks . Some of the researchers have found that predictive neural embedding models have better performance than traditional counting models in calculating word similarity[9, 18], some researchers have published papers that show the traditional counting models have better performancethan than prediction-based neural embedding models on some datasets of specific NLP tasks[19-20].

So far no study has been conducted in this field about comparing the performance of count-based models and prediction neural embedding models, such as LDA, PIM(or other PMI variants) and SGNS which are representative distributional semantic models on the document classification. In this paper, We make a deep research and compare the performance of LDA, SGNS, SPPIM distributional semantic models in text categorization. The Experiment shows that whether the TSVD method or the ISVD method get the poor performance than the PPMI method in the similaraty task in the low frequency range[21]. Especially, the ISVD method get the worst results. The paper analyzed these results as that ISVD removed the latent dimension vectors which have the largest variance, that are no doubt the most important dimensions for fairly low frequent item. Inspired by the paper insight, we propose a WL-SPPIM semantic model based on SPPIM model empirically. Different with SPPIM: 1) We make some changes on one term in the formula that produces the PMI matric, by replacing the $P(c)$ with $p^{3/4}(c)$ ; 2) we reweight the probility of the low frequency word by merging the probability of the high frequency words into the probility of the low frequency words. The experimental results show that, on the document classification, WL-SPPIM yields better results than LDA, SGNS, SPPIM on the accuracy and expansibility.

The rest of the paper is structured as follows: Section 2 introduces the LDA, SGNS, SPPIM semantic models for document classification; Section 3 describes the proposed WL-SPPIM semantic model for document classification; Section 4 presents the experiments; Section 5 discusses the result; and Section 6 conclusions the paper and suggests the future work direction.

## 2 BACKGROUND
### 2.1 LDA-based Semantic Model

LDA is not only a topic model, but also a "text-topic-word" three-level Bayesian model. The text reflects mixes distribution of topic, while the topic demonstrates the probability distribution of words. A document contains many topics, and each word in the document is generated by one topic. As for each word in the document $d = \{w_1, w_2, \ldots w_N\}$, LDA model chooses a topic $\beta_{z_i}$, topic set $\beta$ based on the multinomial distribution $Multi(\theta)$ which is determined by related variable $\theta$, in which $z_i$ represents the number of selected topic, and $\theta$ is consistent with the Dirichlet Distribution $Dir(\alpha)$. After the topic $\beta_{z_i}$ is selected, the word $w_i$ is generated by multionmial distribution $Multi(\beta_{z_i})$ which is defined by $\beta_{z_i}$, the generation of text is summarized by formula (1) - (3).

$$\theta \sim Dir(\alpha) \quad (1)$$
$$z_i \sim Multi(\theta) \quad (2)$$
$$w_i \sim Multi(\beta_{z_i}) \quad (3)$$

Text feature vectors are generated by adopting LDA to extract text features of documents in low dimensional space. These feature vectors can be used for building classification models and classification. This paper extracts text features of LDA according to following steps: 1) set the size of vocabulary $V$ and topic set $\beta$; 2) input all documents of training set (regardless of category) into LDA model, training model parameters $\alpha$ and $\beta$; 3) LDA models obtained from training are used in reasoning to calculate $P(\theta|d)$. In general, the maximum point $\theta$ of $P(\theta|d)$ is used as feature vector of the document, namely $(\theta_{MAP})$, the maximum posteriori probability point $\theta$ : $f(d) = \theta_{MAP}(d) = \arg\max_{\theta} P(\theta|d)$.

### 2.2 SPPIM semantic model

The traditional way to represent words in the distributional approach is to construct a high dimensional sparse matrix $M$, where each row represents a word $w$ and each column represents a context $c$. The value of each matrix cell $M_{ij}$ represents the association between the word $w_i$ and the context $c_j$. A popular measure of thisassociation is pointwise mutual information(PMI)[6]. PMI is calculated empirically like:

$$PMI(w,c) = \log \frac{\#(w,c)|D|}{\#(w) \bullet \#(c)} \quad (4)$$

D is the number of observed word and context pairs.

In terms of negative PMI values may not positively contribute to model performance and sparser matrices are more computationally tractable[2], a sparse and consistent alternative is to use PPMI matric. In PPMI matric, all negative values which are deduced from PMI matric are replaced by 0:

$$PPMI(w,c) = \max(PMI(w,c), 0) \quad (5)$$

SPPMI can be generalized to an additional coutoff parameter $k$, analysising by Levy and Goldberg shows hat SGNS is equivalent to a shifted version of the PPMI method, where all values get shifted by a factor $\log k$ [22]:

$$SPPMI(w,c) = \max(PMI(w,c) - \log k, 0) \quad (6)$$

In the SPPMI model, word distributions are projected into high dimensional sparse space. Distributional semantic models can get better performance, when applying a variety of transformations to the original vector, such as by reweighting to the text and applying the dimensionality reduction technique. So we apply SVD to the SPPMI matrix in order to obtain low-dimensional embeddings. When the SPPIM-SVD matrix produces the word vector, the document vector $v_i$ can be computed also by the simple average pooling approach.

## 2.3 SGNS Semantic Mode

Prediction-based neural embedding models such as the skip-gram model and continuous bag-of-words model have become the standard for word modeling[14, 17]. In the skip-gram model, for a word $w \in V_W$ and a context $c \in V_C$, their embedding vectors are represented as $w \in R^d$ and $c \in R^d$ respectively, where $d$ is the embedding's dimensionality. The embedding vectors of the words in $V_W$ and the context words in $V_C$ compose the word and context embedding matrices $W$ and $C$. It seeks to represent each $w \in V_W$ and each context $c \in V_C$ as d-dimensional vectors $\vec{w}$ and $\vec{c}$, the words will have similar vector representations, if the words are similar.

Levy and Goldberg made an interesting connection between two models: a traditional count-based model on PMI(pointwise mutual information) and a prediction-based neural embedding model, namely the SGNS(skip-gram model with negative sampling)[22]. The interesting result is that the SGNS model is equivalent to a shifted version of the PMI method, where all values get shifted by a factor of $\log k$. Levy and Goldberg described the matrix $M$ that SGNS is factorizing:

$$M_{ij}^{SGNS} = W_i \cdot C_j = \vec{w}_i \cdot \vec{c}_j = PMI(w_i, c_j) - \log k \quad (7)$$

The meaning of the document can be considered a summary of the meaning of all the words in the document. We use the Word2vec tool to generate the word vector first, and then generate the document vector by averaging pooling the word vector. We use $c_{i,j}$ to denote the word vector of the $j$ th word in the document $i$, the document vector $v_i$ can be calculated as: $v_i = \frac{1}{J_i} \sum_{j=1}^{J_i} c_{i,j}$, Where $J_i$ is the number of words in the document.

## 3 WL-SPPIM Semantic model

The Shifted PPMI performs slightly better than SGNS derived vectors on two datasets, but performs poorly than the latter on one dataset of text classification tasks(see section 4).

According to the algorithm illustrated in[17], the negative contexts are sampled according to the formula $p^{3/4}(c) = \frac{\#c^{3/4}}{z}$ instead of the unigram distribution $\frac{\#(c)}{z}$, so we modify the SPPIM formulation:

$$SPPMI(w,c) = \max(PMI(w,c) - \log k^{3/4}, 0) \quad (8)$$

The SPPIM underperforms than the SGNS on the one dataset on text classification task(see section 4), one of the main reason is that SGNS performs weighted matrix factorization. Inspired by this, we propose the WL-SPPIM approach. The probabilities of the low frequency words are generally underestimated in terms of the lack of occurrences in the training datan, and estimating the probabilities of words that absent in corpus is simply impossible. However, these words are often important entity names that should be emphasized, so we have a idea that transform some information from high frequency words to enhance the weight of the low-frequency words. Given a set of words $W = \{x_1, x_2, \ldots, x_m\}$ to be enhanced, for each word $x_i \in W$, a set of words $S_i = \{y_1, y_2, \ldots y_n\}$ that are similar to $x_i$ is manually selected from the training data. The similarity can be defined with semantic. We assume that, for each $y_j \in S_i$, if there exist an n-gram of $y_j$ in the training corpora, the corresponding n-gram of $x_i$ should also have a relative higher probability of appearance. In general, *TF-IDF* method is adopted for the weight design of features[23]. $TFIDF(t,d) = TF(t,d) \times \log_2 \frac{N}{DF(t)}$ is the calculation formula of word frequency-inverse document frequency; $TF(t,d)$ refers to the frequency of word $t$ in document $d$, $N$ is total document number, and $DF(t)$ is the number of other documents containing feature word $t$. But *TF-IDF* lacks the ability to distinguish feature items in different categories. We designed new weight calculation method based on *TF-IDF* technology.

$W(t|c) = TF(t|c) \times \log_2 \frac{N_c}{DF(t|\bar{c})}$ represents the ability of feature item $t$ to distinguish category $c$, $c = c_1, c_2, \ldots c_k$ and $TF(t|c)$ represents the frequency of feature item $t$ in category $c$, $N_c$ is the number of documents in category $c$, and $DF(t|\bar{c})$ represents the the number of documents containing feature item $t$ which are included in the class that not belongs to $c$. The value of $W(t|c)$ is larger, the ability of feature item to distinguish category $c$ is the stronger. $W(t|c)$ measures the ability of feature word to distinguish categories from global perspective.

Then we adjust the weight of the low-frequency according to the following equation:

$$w_{x_i} = w_{y_j} + \ln\left(\frac{W(x_i|c)}{\sqrt{\sum_{c_i \in c} W(x_i|c_i)^2}}\right) \quad (9)$$

Where $x_i$ presents the low-frequency word.

$W(x_i | c)$ indicates the ability of $x_i$ to distinguish category $c_i$, $TF(x_i | c_i)$ represents the frequency of $x_i$ in category $c_i$, $N_{c_i}$ is the number of documents in category $c_i$, $DF(x_i | \overline{c_i})$ is the number of documents containing word $x_i$ which are included in the class that not belongs to $c$. In addition, a threshold is set to guarantee that $\frac{W(x_i | c)}{\sqrt{\sum_{c_i \in c} W(x_i | c_i)^2}}$ is not less than this threshold, if the value is too small to enhance the recognition performance, this threshold will be assigned to it.

# 4 EXPERIMENT

## 4.1 Database and configurations

We experiment all models on English corpus, preprocessing by removing single characters and alphabetic characters, and converting letters to lowercase, stemming word roots, sentence splitting, and tokenization. The experiments were conducted with three datasets:20 Newsgroups dataset that was originally collected by Ken Lang[1], WebKB dataset that was collected by the project of the CMU[2], and Reuters dataset published by DavidD Lewis[3]. Table 3 lists the statistics of these three datasets.

20Newsgroups corpus contain 20 categories of English news text, which contains a total of 18846 Document. In addition,in order to improve the reliability of experiment, all repeat documents and some news heads are removed, which left 11293 and 7528 documents to the training data and testing data. Original WebKB corpus contains about 8,300 English websites, which can be divided into 7 categories. Selecting the most commonly used 4 major categories, including student, faculty, course and project categories, this text subset is called WebKB-4. In addition, in order to improve the reliability of experiment, some repeat documents are removed, which left 2756 and 1375 documents to the training set and testing set. The Reuters dataset is a collection of Reuters Newswire in 1987. We have 8 of the 10 most frequent classes and 52 of the original 90, namely R8 and R52. R52 is a totally 4M corpus which contains 3M text (6532 documents) for training, and 1M text(2568 documents) for testing.

We use the word2vec tool provided by Googleused to train the skip-gram word vector[4]. LDA training and learning tools invented by Blei is used to generate document vector[5]. We conducted text classification experiments on three classifiers, including Bayesian, KNN, and SVM which all are build by the scikit-learn tool[6].

TABLE 1
DATASETS STATISTICS

| Datasets | Train | Test | Word | Stop |
|---|---|---|---|---|
| 20ng | 11293 | 7528 | 93864 | no |
| R52 | 6532 | 2568 | 26284 | no |
| WebKB | 2756 | 1375 | 7770 | no |

## 4.2 Experiments

The first experiment investigates the performances of document classification methods LDA, SGNS, SPPIM and WL-SPPIM when the dimensionality of document vector is different. In LDA method, dimensionality denotes the number of topic in topic set $\beta$; in SGNS, SPPIM and WL-SPPIM, dimensionality refers to the dimension of space. The document classification performance of document vectors generated by the above four semantic models in 20News, Reuters52 and WebKB datasets are tested. Figure 1-3 shows classification performance of text features generated by above 4 models in 20News, R52 and WebKB, respectively, the abscissa refers to the dimension of document vector and ordinate the accuracy of document classification on the three datasets.

SGNS has a natural hyper parameter $k$, which is the number of negative samples, which affects the value that SGNS is trying to optimize for each $(w,c): PMI(w,c) - \log k$. The shift caused by $k > 1$ can be applied to distributional methods through shifted PPMI. SPPMI achieve good performance with lower values of $k$, what is due to the fact that only positive values are retained, and high values of $k$ may cause too much loss of information[22]. Empirically, we take experiment with values of $k = 5$.

At first, We experiment these models with different classifiers on 20newsgroups, and the results about classification accuracy are depicted in Table 2. We conduct this experiment with 50-dimension on the word vector in the SGNS and the topics in the LDA, then the dimensions of the document vectors will be 50-dimension with four approaches. We observe that KNN method outperforms the other two, when the three classifiers are compared, which the result is showed in Table 2. Nevertheless, KNN is a nonparametric method, which is not suitable for large corpus, so we choose SVM as the classifier which will be used in the following experiments.

TABLE 2
CLASSIFICATION WITH DIFFERENT CLASSIFIERS ON 20NEWSGROUPS

| Classifier | LDA | SNGS(W2V) | SPPIM | WL-SPPIM |
|---|---|---|---|---|
| NB | 61.47% | 63.69% | 62.71% | 66.33% |
| KNN | 78.21% | 75.15% | 72.10% | 77.63% |
| SVM | 71.67% | 75.15% | 72.94% | 77.98% |

---

[1] http://www.cs.cmu.edu/afs/cs.cmu.edu/project/theo-20/www/data/news20.html
[2] http://web.ist.ult.pt/~acardoso
[3] https://kdd.ics.uci.edu/databases/reuters21578/reuters21578.html
[4] https://code.google.com/p/word2vec
[5] http://www.cs.princeton.edu/~blei/lda-c
[6] http://scikit-learn.org/dev/modules/svm.html#svm

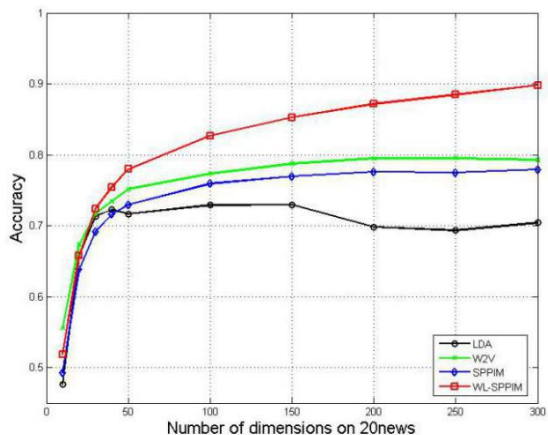

Fig. 1.Performance on 20-Newsgroups

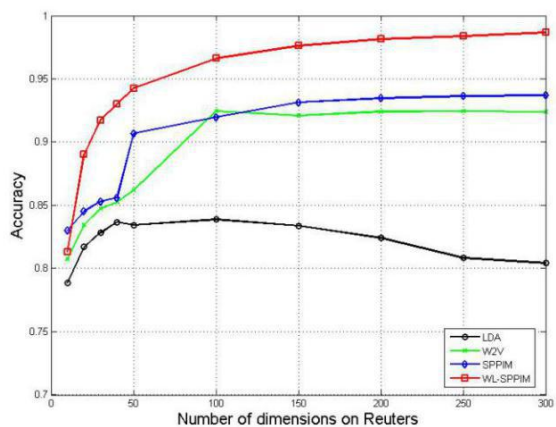

Fig. 2. Performance on Reuters52

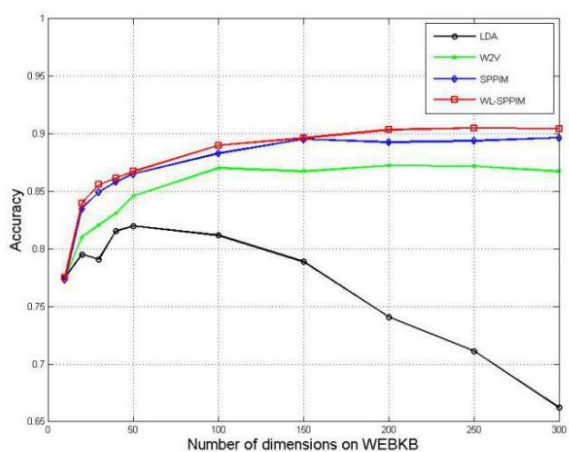

Fig. 3. Performance on WebKB

The second experiment compared the performances of the four models on document classification of different complexity. Figure 4 shows that text feature dimension is set at 200 dimensions, and classification complexity is increased from two-level classification to 20-level classification in 20Newsgroups dataset, and from two-level classification to 4-level classification in WEBKB dataset to test the performance of the classification task complexity of the four models in each classification task. For each classification task, LDA, SPPIM and WL-SPPIM should retrain parameters in corresponding training corpus, while word vector generated by word2vector tools does not depend on specific text field, so SGNS model does not retrain parameters. In the Figure 1-3, the abscissa represents different complexity on document classification, and ordinate the accuracy of classification.

## 5 EXPERIMENTAL RESULTS AND ANALYSIS

We evaluate LDA, SGNS, SPPIM, WL-SPPIM on the text classification task. The comparison of these four models in terms of the text classification accuracy is seen in Fig 1-Fig3 , the dimensionality of the document vectors varies from 10 to 300 in our experiment . Fig1-Fig3 shows the WL-SPPIM model achieves its best performs on all the datasets, WL-SPPIM outperforms SPPIM consistently. And the SPPIM is subtly better than SGNS on R52 and WebKB datasets, the accuracy of SPPIM and SGNS are very close on 20newsgroups datasets. WL-SPPIM significantly outperforms SPPIM and SGNS, verifying that re-weighting the weight of the word vector could improve the accuracy of the document classification. Given the similar performance of SPPIM and SGNS in the text classification task, the superiority of WL-SPPIM is obvious. The accuracy of LDA is much lower than that of other methods, which may be because LDA performs better with long documents[24]. It can be learned that the accuracy of the four models in classification improves with the increase of document vector dimensions, indicating document vector dimension is larger, the classification accuracy is better. However, the classification performance of LDA in WebKB data decreases sharply with the increase of feature dimensions, which is attributed to the lack of long text training corpus in the field, difficult to meet the optimization task of LDA n high dimensional vector model.

As to the effects of different category, one can observe that the performance of WL-SPPIM remains stable with different categories from Figure 4  The accuracy of LDA is much lower than that of other methods, which may be due to uneven distribution of topic vector in various categories and thereby leading to differences in various categories as LDA is greatly influenced by model initial value in the establishment of topic model. As the generation of document vector by SGNS combination involves the process of mean smoothing, so it is less affected by categories changing and the classification performance is stable. Similarly, reweighting the low frequency words in WL-SPPIM model is equivalent to smoothing techniques, so multi-classification experiments shows better classification performance and higher stability.

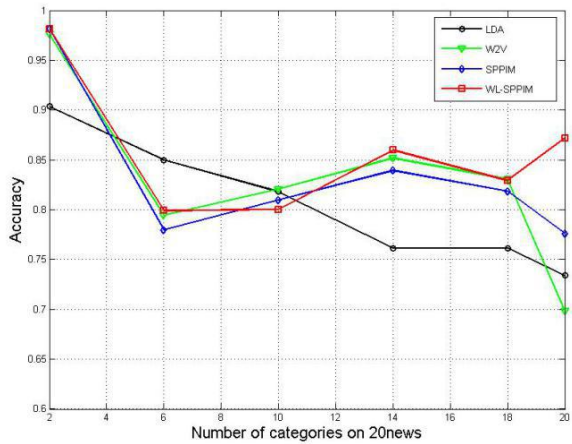

Fig. 4. Experimental results with various numbers of category on 20-newsgroups

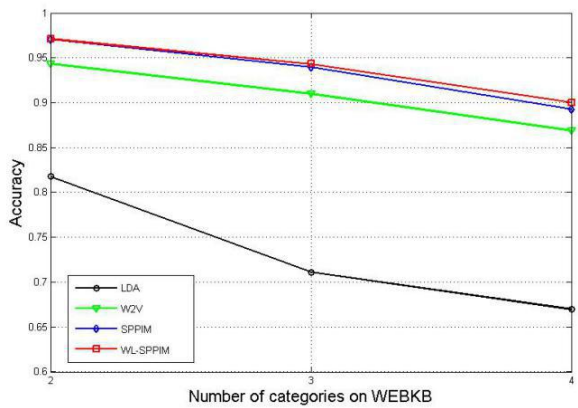

Fig. 5. Experimental results with various numbers of category on WebKB

From the results revealed in Fig4 and Fig5, we can observe that the proposed method produce good accuracy on the 20newsgroups dataset and R52 dataset, but not satisfied on the WebKB dataset. Improving the weight of low-frequency words does not enhance the classification performance of WL-SPPIM model in the WebKB dataset, which may be because the WebKB dataset is relatively small, so low frequency and high frequency words meeting the requirements of similar-pair are very few. This result confirms the effectiveness and high performance of proposed method. This method could be applied on semantic tasks in specific domain where low-frequency words need to be enhanced.

The experimental results reveal that our count-based model achieve better performance than prediction-based neural embedding model on the document classification task. Furthermore, these experiments not only verify the classification performance of WL-SPPIM, but also the generalization ability of WL-SPPIM.

## 6 CONCLUSION

In this paper, we verify the discovery in[20, 25] on the text classification task, what the traditional count-based models achieve better performance than prediction-based neural embedding models on some specific NLP tasks. Furthermore, we propose a method based on WL-SPPIM semantic model for the document classification, and investigate the characteristics of LDA, SGNS, SPPIM and WL-SPPIM sematic models on the semantic source and semantic generation method in detail, and analyze the advantages of WL-SPPIM semantic model on the document classification. Experiments shows that the WL-SPPIM model is significantly superior than other three methods in classification accuracy, complexity, scalability on text classification tasks. In future work we will investigate distributions semantic clustering for WL-SPPIM model.

## ACKNOWLEDGMENT

This research was supported by the National Science Foundation of China (NSFC) under the project No.61672488, Ministry of science and Technology "Key technologies of educational cloud " project No.2013BAH72B01, Chongqing Science and Technology Commission project No.cstc2013jjys4000. It was also partially supported by the Magicalthink Ltd.